\documentclass[10pt,twocolumn,letterpaper]{article}

\usepackage{cvpr} 
\usepackage[dvipsnames]{xcolor}

\definecolor{cvprblue}{rgb}{0.21,0.49,0.74}
\usepackage[pagebackref,breaklinks,colorlinks,citecolor=cvprblue]{hyperref}
\usepackage[symbol]{footmisc}

\def\paperID{40} 
\def\confName{ECCV}
\def\confYear{2024}

\title{Counting Fish with Temporal Representations of Sonar Video}

\author{
    Kai~Van~Brunt$^{1, \dagger}$, Justin~Kay$^1$, Timm~Haucke$^1$, Pietro~Perona$^{2}$, Grant~Van~Horn$^{3}$, Sara~Beery$^{1}$ \\
    $^{1}$MIT, $^{2}$Caltech, $^{3}$UMass Amherst
}

\begin{document}
\maketitle
\footnotetext[2]{Correspondence to kav@mit.edu}

\begin{abstract}
Accurate estimates of salmon \textit{escapement}---the number of fish migrating upstream to spawn---are key data for conservation and fishery management. Existing methods for salmon counting using high-resolution imaging sonar hardware are non-invasive and compatible with computer vision processing. Prior work in this area has utilized object detection and tracking based methods for automated salmon counting. However, these techniques remain inaccessible to many sonar deployment sites due to limited compute and connectivity in the field. We propose an alternative lightweight computer vision method for fish counting based on analyzing \textit{echograms}---temporal representations that compress several hundred frames of imaging sonar video into a single image. We predict upstream and downstream counts within 200-frame time windows directly from echograms using a ResNet-18 model, and propose a set of domain-specific image augmentations and a weakly-supervised training protocol to further improve results. We achieve a count error of 23\% on representative data from the Kenai River in Alaska, demonstrating the feasibility of our approach.

\end{abstract}    
\section{Introduction}
\label{sec:intro}






Accurate salmon population monitoring enables 
data-driven fishery management and conservation. In particular, fishery managers and conservationists are interested in salmon \textit{escapement}: the abundance of migrating salmon returning from the sea that successfully spawn. Several methods exist for monitoring migrating salmon (see \cref{sec:related}). 
Sonar-based monitoring has recently grown in popularity due to its non-invasive nature and ability to collect data at high temporal resolution under a variety of conditions.
However, sonar cameras produce large amounts of data---in some cases over 30GB of data a day~\cite{AlaskaSonar}---and 
reviewing this data
is time-intensive for technicians, with no existing alternative that generalizes across sites.

Computer vision has the potential to more efficiently and accurately analyze sonar video for escapement monitoring. Prior work has introduced automated approaches based on object detection and multi-object tracking~\cite{kay2022caltechfishcountingdataset}. These approaches achieve counting errors of under 10\%; however, they rely upon processing each video frame independently with deep networks (\eg YOLOv5m with 21.2M params~\cite{kay2022caltechfishcountingdataset}), making them currently unsuitable for deployment in locations with limited compute and connectivity.

\begin{figure*}[t]
  \centering
  \includegraphics[width=0.6\linewidth]{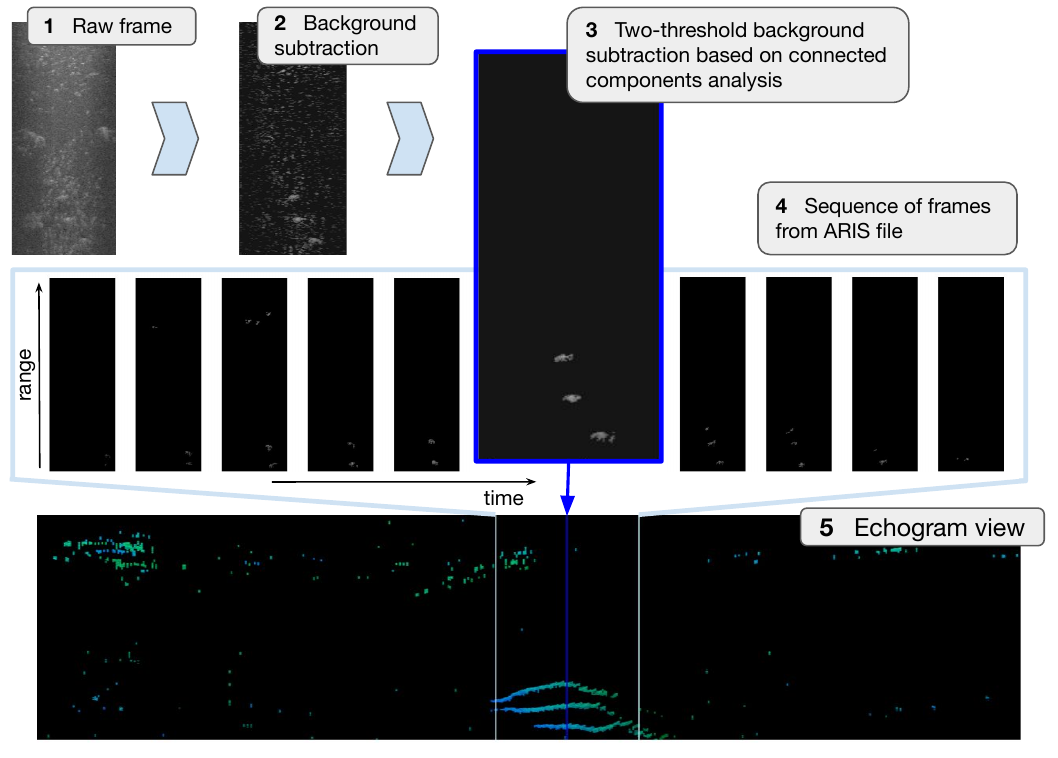}
  \vspace{-10pt}
  \caption{Clockwise: 1) a frame of the raw ARIS file; 2) the same frame after applying background subtraction with a minimum positive threshold on each pixel intensity of $\alpha_0=10$ above the mean frame; 3) the same frame after applying connected components analysis, applying background subtraction with a threshold of $\alpha_2=127$ outside the largest connected components and a threshold of $\alpha_1=35$ inside the largest connected components. 4) Selected frames from a time range in an ARIS file and 5) the same length of time displayed in echogram view, where the color corresponds to the lateral position of the brightest pixel.
  }
  \vspace{-10pt}
  \label{fig:echogram}
\end{figure*}

In this paper, we explore an alternative approach to automated salmon counting in sonar video that harnesses a temporal representation called an \textit{echogram}. Echograms compress a multi-beam sonar video into a 2D image (see \cref{fig:echogram}). The $x$-axis of an echogram represents time. At each $x$-value, a column vector represents a compressed view of an entire frame of video. In this column vector, 
the pixel intensity at each $y$-value corresponds to the maximum intensity across all sonar beams at the corresponding range. If fish are present, this will result in a noticeable visual signature. Sonar technicians use these echogram visualizations during data review, 
both to identify temporal regions of interest 
and to cross-check challenging counts. 

We propose and evaluate the feasibility of a method for analyzing echograms with computer vision to directly predict fish counts, providing a low-compute alternative to object detection and tracking pipelines. Our method takes as input a 200px wide echogram image and predicts the number of fish moving upstream or downstream during the corresponding timeframe, thus requiring only a single forward pass every 200 frames through a lightweight backbone (\eg a ResNet-18 with 11.7M params~\cite{he2016deep}) to compute counts. We further propose a set of domain-specific image augmentations as well as a weakly-supervised training protocol that incorporates annotations generated ahead of time by an object detector and tracker. 

Our initial model achieves counting error rates of 23\% on a validation set that is in-distribution with respect to the training set and 30.7\% on an out-of-distribution test set, nearly matching initial proofs of concept for much more computationally expense tracking-by-detection approaches~\cite{kulits2020automated}. We perform quantitative and qualitative analyses to identify challenges in echogram-based approaches as well as promising areas for future work.

\section{Related work}
\label{sec:related}

\noindent
\textbf{Salmon escapement monitoring.} Several methods exist for monitoring salmon escapement, including weirs, counting towers, and various sonar hardware. We include a broader overview of these methodologies in the supplemental material.
In this paper we focus on a relatively new generation of sonar hardware known as \textit{imaging sonar} that produce high-resolution videos using multi-beam acoustic hardware. Imaging sonar can be accurately analyzed to count and measure salmon by both human technicians~\cite{AlaskaSonar} as well as computer vision systems~\cite{kay2022caltechfishcountingdataset}.

\noindent
\textbf{Computer vision for salmon monitoring.}
Existing computer vision approaches to salmon counting utilize tracking-by-detection to first perform object detection on individual frames and then link together predicted bounding boxes into trajectories~\cite{kay2022caltechfishcountingdataset,ouis2023yolo}.
Once these tracks are determined, different heuristics may be used to determine fish counts. While such approaches produce accurate counts when the training river and testing river are the same, they struggle on out-of-distribution test data sourced from \eg different rivers or different environmental conditions than the training data~\cite{kay2022caltechfishcountingdataset,kay2024align,kay2023unsupervised}. Another key challenge for real-world deployment of these models is their compute requirements, as fish counting technicians are often stationed in remote locations with only consumer laptops; on such hardware, even efficient object detection and tracking techniques like YOLO~\cite{redmon2016you} and SORT~\cite{bewley2016simple} are a severe processing bottleneck. Our method aims to enable more efficient inference by bypassing frame-by-frame video analysis through compressed temporal representations called echograms.

\section{Method}

\subsection{Echogram generation}
\label{sec:echogram}

We begin with sonar video files in the ARIS format~\cite{soundmetrics_ARIS}. Each file represents 10-20 minutes of continuous sonar footage which may or may not contain fish. At each range (radial distance from the sonar camera) a certain angular span is sampled, 
outputting a pixel intensity corresponding to the strength of the echo received. 

To generate the echogram, we apply successive iterations of background subtraction to each frame of the ARIS file, as in \cref{fig:echogram}. In each application of background subtraction, after taking the mean frame across the ARIS file, only the pixels of each frame exceeding a threshold value $\alpha$ above the corresponding pixel of the mean frame are kept. 

First we apply background subtraction to the raw frame with a low threshold value $\alpha_0$. Then OpenCV's ConnectedComponentsWithStats function obtains all connected components in the new image which are larger than a size threshold scaled by range. Finally, we apply background subtraction once more with a threshold value $\alpha_1$ within these components and $\alpha_2$ outside these components such that $\alpha_0 < \alpha_1 < \alpha_2$.
The values of each threshold are tuned by trial and error until a qualitatively acceptable echogram is produced, and we use those same parameters for all data.


This version of the clip, with each frame cleaned of background noise, is used for echogram generation. Each frame of shape (number of samples along range) $\times$ (number of beams) is collapsed into a column of height (number of samples along range), each pixel of the column corresponding to the maximum intensity at that range of the various beams. A second image channel stores the lateral position of that maximum intensity point, normalized between 0 and 1. Concatenating these columns together gives the full 2D echogram, of shape (number of samples along range) $\times$ (number of frames in video).

\subsection{Computer vision model}

We train a computer vision model in the PyTorch Lightning machine learning framework to predict left and right counts for echogram images. We finetune a ResNet18 model pre-trained on ImageNet with a final fully connected layer that contains two outputs corresponding to left and right counts. We use a ReLU activation function after the final layer (since counts must be non-negative) and optimize for mean squared error. 
We use an input size of 200px by 800px, learning rate of 1e-5 using Adam optimization, batch size of 256, and train for a maximum of 100 epochs on a single NVIDIA A100 GPU with early stopping based on KL-val (see \cref{sec:data}) performance.
\section{Dataset and metrics}

\subsection{Data collection and annotation}
\label{sec:data}

We generate echograms for the Caltech Fish Counting dataset (CFC) from \cite{kay2022caltechfishcountingdataset}. We use the default training and validation sets, ``KL-train'' and ``KL-val'' from the \textbf{l}eft bank of the \textbf{K}enai River in Alaska, and we also test on one out-of-distribution test set, ``KR'' from the Kenai \textbf{r}ight bank. 
In total, this gives us 481 KL-train images, 66 KL-val images, and 406 KR test images. We refer to the ground truth count labels for CFC as \textbf{strong labels} in our experiments.

We also generate additional \textbf{weak labels} on a set of previously-unlabeled ARIS files collected from the same camera locations as the KL-train and KL-val sets. These weak labels are generated by the public detector and tracker pipeline released with CFC~\cite{kay2022caltechfishcountingdataset}. We label counts in the same way as \cite{kay2022caltechfishcountingdataset}: a fish whose trajectory start and end are on opposite sides of a vertical line drawn through the center of the frame is counted as either an\textit{left} or \textit{right} traveling fish, based on the relative start and end points of the trajectory. We ensure there is no overlap between the KL validation set and the detector-tracker annotated training or validation set.
In total, we generated weak labels using this pipeline for 33,437 images from the KL location.

There is a large imbalance between leftward and rightward moving fish, since the data is collected to monitor salmon migrating upstream. We orient all clips such that right-moving fish travel upstream and left-moving fish travel downstream, to make the model invariant to the physical upstream direction.

\subsection{Metrics}

To evaluate model performance we use the normalized Mean Absolute Error (nMAE) as in prior work~\cite{kay2022caltechfishcountingdataset}:

\vspace{-5pt}
\begin{equation}
    \text{nMAE}=\frac{\sum_{i=0}^{N} E_i}{\sum_{i=0}^{N}\hat{z}_i}
\end{equation}

\noindent
where $N$ is the number of clips, $\hat{z}_i$ is the target number of counts on the $i$th clip, and the error $E_i$ is the sum of absolute errors on left and right counts on the $i$th clip.
We also report nMAE for left and right counts separately. 

\section{Experiments}



\begin{figure}[t]
  \centering
  \includegraphics[width=0.95\linewidth]{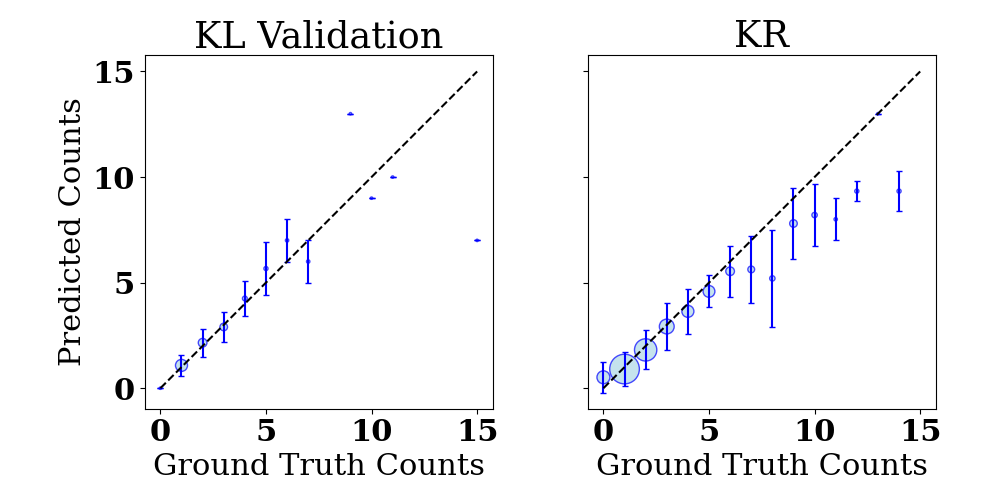}
  \vspace{-5pt}
  \caption{\footnotesize 
 Mean and standard deviation of total predicted counts vs total ground truth counts per clip on the KL-val and KR test sets. Size of the dot corresponds to the number of images with the associated ground truth count. The model systematically predicts lower counts than ground truth for KR clips with large numbers of fish, where tracks of distinct fish may overlap and become difficult to distinguish on the echogram.
  }
  \vspace{-10pt}
  \label{fig:preds_gt}
\end{figure}

Our best-performing model achieves an overall nMAE on KL-val of 23\% and 30.7\% on KR (\cref{tab:datasets}).
The count error on downstream-moving fish is especially high (\cref{tab:datasets}), due partly to an extreme class imbalance between downstream- and upstream-moving fish in all training sets. In addition, the model systematically predicts lower counts than ground truth in clips with large numbers of fish (\cref{fig:preds_gt}), where separate tracks on an echogram may overlap and become difficult to distinguish.

These error rates are higher than state-of-the-art detector-tracker pipelines for salmon counting: Kay et al.~\cite{kay2022caltechfishcountingdataset} achieved 4.9\% error on KL-val and 11.8\% on KR using a YOLOv5m detector, and reduced these errors to 3.3\% error on KL-val and 3.7\% error on KR using a more complex input representation. However, our results are comparable to initial results published by the same team~\cite{kulits2020automated} that reported counting error rates of 19.3\%, indicating the potential to improve our results in future work.

Our experiments in \cref{tab:datasets}, \cref{tab:echogram-params}, and \cref{tab:dataaug} demonstrate that our model's performance is improved by incorporating both weak and strong labels during training, tuning the echogram generation parameters, and by applying domain-specific data augmentations during training. We ablate these contributions next.

\subsection{Training data}
\vspace{-10pt}
\begin{table}[h]
\centering
\resizebox{\linewidth}{!}{
\begin{tabular}{lcccccc}
\toprule
 & \multicolumn{3}{c}{\textbf{KL-val nMAE (\%)} $\downarrow$} & \multicolumn{3}{c}{\textbf{KR nMAE (\%)} $\downarrow$} \\
\cmidrule(lr){2-4}\cmidrule(lr){5-7}
\textbf{Training set} & Total & Down & Up & Total & Down & Up \\
\midrule
KL-train {\footnotesize(\cite{kay2022caltechfishcountingdataset}, 481 imgs)} & 42.1 & 100.0 & 39.4 & 61.0 & 102.7 & 57.9 \\
KL-weak {\footnotesize(Ours, 33k imgs)} & 44.3 & 112.5 & 41.1 & 34.8 & 96 & 30.3 \\
\textbf{KL-train + KL-weak} & \textbf{23.0} & \textbf{37.5} & \textbf{22.3} & \textbf{30.7} & \textbf{96.0} & \textbf{25.8} \\
\bottomrule
\end{tabular}
} 
\vspace{-5pt}
\caption{\footnotesize 
 Dataset choice vs performance on KL-val and KR, split by downstream (``down'') and upstream (``up'') moving fish.  Training with strong and weak labels improves over both a small dataset of strong labels only, and a large, diverse dataset of weak labels only.}
 \vspace{-5pt}
\label{tab:datasets}
\end{table}

We train the model on three different datasets: one composed of weak labels only; one composed of strong labels only; and one composed of a mixture of all weak labels and strong labels. In \cref{tab:datasets}, the drastic difference between nMAE for the model trained only on strong labels vs the mixture of weak and strong labels (about a 20\% improvement for KL-val and 30\% improvement for KR) indicates that training on a large, diverse dataset improves the model despite the potential inaccuracies present in the weak labels. The inclusion of strong labels, which make up less than 2\% of the total dataset size, also significantly improves model performance compared to the model trained on weak labels only, especially on the in-distribution test set (KL-val). 

\subsection{Echogram generation parameters}

\vspace{-10pt}

\begin{table}[h]
    \centering
    \footnotesize
    \begin{tabular}{ccccc}
        \toprule
        \multicolumn{4}{c}{\textbf{Echogram params}} & \textbf{nMAE (\%) $\downarrow$} \\
        \cmidrule(lr){1-4} \cmidrule(lr){5-5}
        $\alpha_0$ & $\alpha_1$ & $\alpha_2$ & \textit{size\_thresh} & KL-val \\
        \midrule
        0 & 0 & 0 & 0 & 84.7 \\
        20 & 0 & 0 & 0 & 36.6 \\
        \textbf{20} & \textbf{40} & \textbf{60} & \textbf{100} & \textbf{23.0}  \\
        20 & 40 & 100 & 120 & 37.2 \\
        \bottomrule
    \end{tabular}
    \vspace{-5pt}
    \caption{\footnotesize Echogram generation parameters vs performance on KL-val and KR for models trained and validated on a mixture of weak and strong labels. The model performs best at some intermediate setting where a balance is achieved between filtering out background noise and preserving information about fish tracks.}
    \vspace{-5pt}
    \label{tab:echogram-params}
\end{table}

When generating the echogram slices used as input and test data for the model, thresholds for initial background subtraction, secondary background subtraction, and filtering based on size can be tuned. Higher thresholds lead to information loss but also produce a cleaner, less noisy signal for the model. Testing different sets of thresholds as in \cref{tab:echogram-params} shows that an intermediate setting is ideal: the model benefits from some filtering of noise but is negatively affected by cutting background signal too aggressively.


\subsection{Data augmentations}


We explore various data augmentation strategies, informed by the specifics of the echogram image domain, in \cref{tab:dataaug}.

\noindent
\textbf{Vertical flip.} Flipping the entire image across the horizontal axis improves nMAE for all model setups.

\noindent
\textbf{Naive horizontal flip.} Regardless of the set of labels the model is trained on, nMAE worsens when a horizontal flip augmentation is applied, which flips the entire image. In all training and validation sets, a class imbalance between upstream- and downstream-traveling fish exists: during spawn season, many more fish are swimming upstream than downstream. In KL-val, 175 fish are swimming upstream while only 8 are swimming downstream. In addition, the upstream and downstream motion patterns of fish are different due to the direction of the river current. This naive horizontal flip augmentation thus both obscures the true distribution of upstream vs. downstream counts and does not accurately capture the motion of the fish in the opposite direction, suggesting that a different method is needed to robustly classify downstream-swimming fish.

\noindent
\textbf{Realistic horizontal flip.} This transformation reflects the image across the horizontal axis and then inverts the lateral position channel to match the original, pre-reflection direction of fish motion. This improves nMAE across all model setups but does not improve the left-right class imbalance.

\noindent
\textbf{Superposition.} Two echograms are superposed, displaying at each point the intensity and color of the brightest pixel; the target counts are added together. This augmentation has mixed effects on performance, modestly improving training on weak labels while worsening training on strong labels.

\begin{table}[h]
\centering
\resizebox{\linewidth}{!}{
\begin{tabular}{cccccc}
\toprule
\multicolumn{4}{c}{\textbf{Data augmentations}}  & \multicolumn{2}{c}{\textbf{Train set}} \\
\cmidrule(lr){1-4}\cmidrule(lr){5-6}
V. flip & H. flip & Superpos. & Realistic h. flip & KL-train & KL-weak \\
\midrule
& & & & 48.1 & 64.5 \\
\textbullet & & & & 43.2 & 49.2 \\
\textbullet & \textbullet & & & 111.5 & 53.6 \\
\textbullet & \textbullet & \textbullet & & 129 & 49.7 \\
\textbullet & & \textbullet & \textbullet & 74.3 & \textbf{37.2} \\
\textbullet & & & \textbullet & \textbf{39.3} & 44.3 \\
\bottomrule
\end{tabular}
} 
\vspace{-5pt}
\caption{\footnotesize 
 KL-val nMAE using either KL-train or KL-weak and ablating data augmentations, using cleanest echogram generation settings. A naive horizontal flip augmentation worsens the performance of the model trained on either dataset, while vertical flip and our domain-specific realistic horizontal flip improve results.}
 \vspace{-12pt}
\label{tab:dataaug}
\end{table}

\section{Conclusions}

We introduce a new method for salmon population monitoring based on an \textit{echogram}, a 2D representation of an entire sonar video clip, that is more computationally efficient than existing methods for determining fish counts which are applied to individual frames of a sonar video. Our initial results are promising: a lightweight ResNet-18 model achieves significant reductions in nMAE which bring us to count errors comparable to proofs of concept of more expensive models, through appropriate dataset selection, echogram generation, and data augmentation.

Future evaluations and iterations on this model should address the class imbalance between upstream- and downstream-moving fish, develop a larger and more diverse validation set, and fine-tune the echogram generation and data augmentation procedures. 
\section*{Acknowledgements}

This material is based upon work supported by: the MIT Climate and Sustainability Consortium Scholars Program, MIT J-WAFS seed grant \#2040131, National Science Foundation award \#2330423, and Caltech Resnick Sustainability Institute Impact Grant ``Continuous, accurate and cost-effective counting of migrating salmon for conservation and fishery management in the Pacific Northwest.'' Thanks to Erik Young and Suzanne Stathatos for input and discussions, and Bill Hanot for initial conversations on echogram generation.
{
    \small
    \bibliographystyle{ieeenat_fullname}
    \bibliography{main}
}

\clearpage
\setcounter{page}{1}
\maketitlesupplementary

\section{Salmon monitoring additional information}

Several methods exist for monitoring salmonid escapement, each with their own trade-offs. In narrow, shallow streams, constructing weirs (fences with a gate controlled by a technician) allows technicians to count salmon one by one as they pass through. Counting towers---structures built on a stream bank which give technicians an unobstructed, overhead view of their side of the stream---serve a similar purpose. To sample a greater number of streams in a certain area, fisheries also conduct manual aerial surveys from aircraft. In large or turbid rivers, however, these methods fail to produce reliable results. In addition, these techniques are often only applied to a subset of the duration of fish passage---\eg the first ten minutes of the hour, every hour---with results extrapolated to a 24 hour period~\cite{AlaskaNonSonar, Key2017}. 

For several decades, split-beam sonar has been used in rivers in Alaska and the Pacific Northwest to monitor salmon populations. It functions even at night or in turbid waters and provides bank-to-bank coverage~\cite{Key2017}.

\begin{figure}[t]
  \centering
  \includegraphics[width=0.4\linewidth]{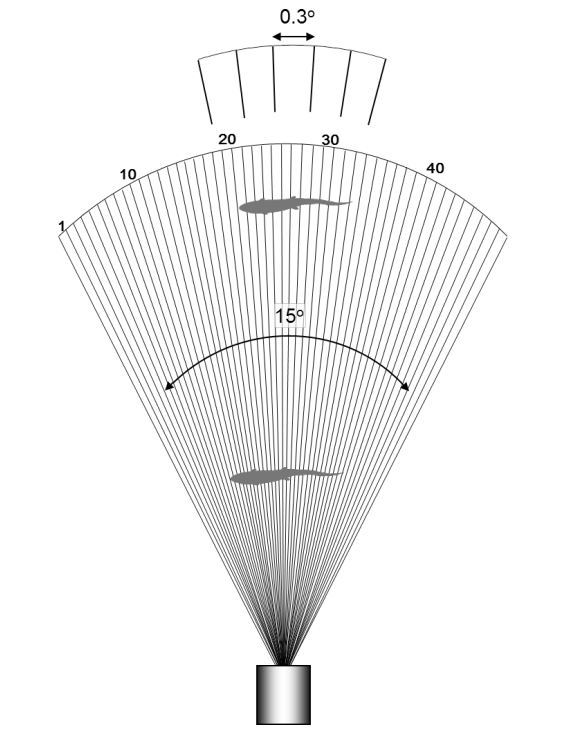}
  \includegraphics[width=0.5\linewidth]{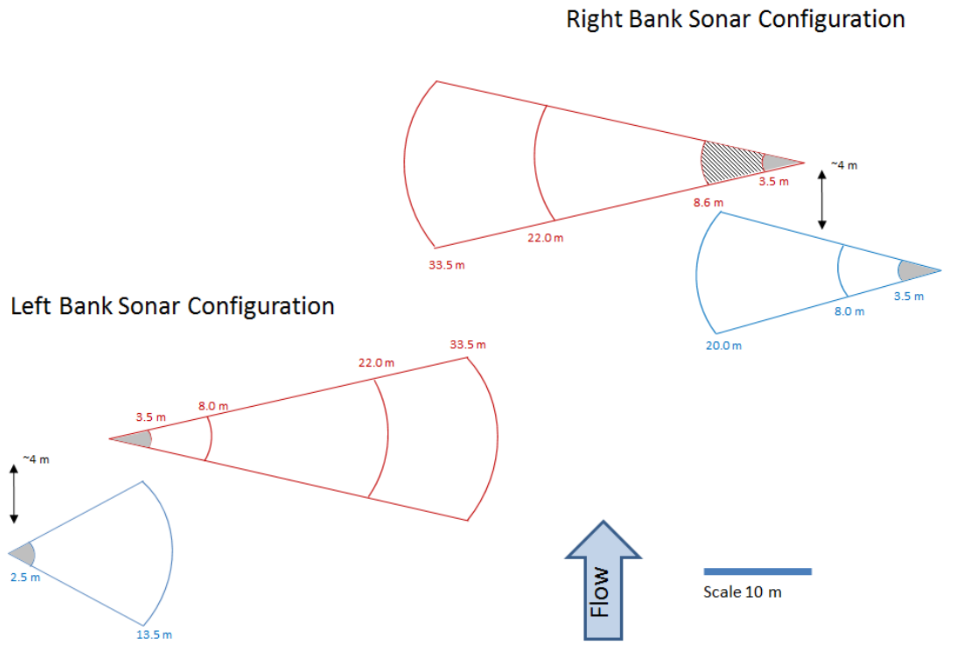}
  \caption{Left: depiction of the horizontal plane of a multi-beam sonar configuration; right: camera placement on the left and right banks of the Kenai river in Alaska~\cite{Key2017}.
  }
  \label{fig:sonar}
\end{figure}

Recently, imaging sonar with multiple beams (Fig \ref{fig:sonar}) has become more popular since it provides a higher-resolution view of the river. It is now the most popular method in rivers such as the Kenai in Alaska or the Eel in California. 
Upon manual review and under careful placement of sonar cameras to e.g. minimize blind spots, imaging sonar can produce counts with high precision ($<$3\%) and with similar accuracy to weir-based counting methods~\cite{holmes2006accuracy}. Fish lengths can also be determined with high precision~\cite{COOK201959}. Both of these findings are complicated by the presence of excessive debris or the passage of high-density schools, which can confuse or obstruct individual fish.

\subsection{Sonar data review process}

Sonar-based monitoring comes with its own challenges: in some cases the cameras can produce over 30GB of data a day, which need to be manually reviewed. Depending on the specific needs of the river (\eg conservation vs recreational fishing management), sonar footage can be reviewed in bulk after the season or almost in real time. This process of manual review---finding portions with fish, and measuring fish lengths so as to filter by species---is time-intensive for the technicians~\cite{AlaskaSonar}.

\begin{figure}[b]
  \centering
  \includegraphics[width=0.7\linewidth]{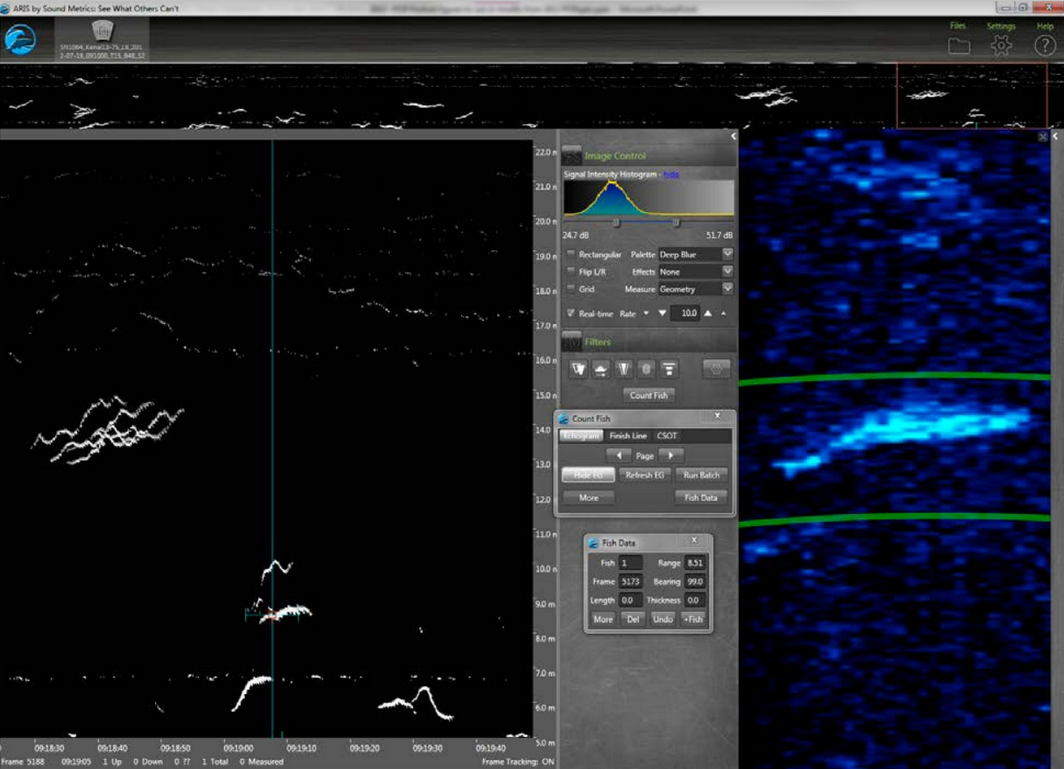}
  \caption{ARIS display software used by sonar technicians showing the echogram view and corresponding frame in sonar video~\cite{Key2017}.
  }
  \label{fig:aris}
\end{figure}

One step that helps streamline the process is the usage of an \textit{echogram}, a 2D representation of the entire length of the clip: each column of an echogram represents one frame of a clip, with pixel intensity corresponding to the maximum intensity across all sonar beams at that range. Technicians use this representation, generated by the proprietary DIDSON or ARIS software which processes camera footage, to identify parts of the clip that are worth watching in full to perform counts and length measurements (Fig \ref{fig:aris}). Still, especially in rivers with high fish passage, obtaining full counts even in a subset of the data stream is a time-intensive and error-prone process.

\section{Data preprocessing}

All images in the training, validation, and test sets are subject to a sequence of transformations to standardize the input format. The available transformations are described below in order of application.

\begin{enumerate}
    \item \textbf{Shift and rescale all channels.} Shifts all pixel values (initially lying between 0 and 1) down by 0.5 and divides the result by 0.25.
    \item \textbf{Resize to 200 by 800 pixels.}
\end{enumerate}

\end{document}